\documentclass[sigconf]{acmart}

\AtBeginDocument{%
  }

\usepackage{subfigure}
\usepackage{xurl}
\usepackage{multirow}
\usepackage{enumitem}
\usepackage{algorithm}
\usepackage[noend]{algpseudocode}

\begin{document}

\title{CarbonScaling: Extending Neural Scaling Laws for Carbon Footprint in Large Language Models}

\author{Lei Jiang}
\affiliation{%
   \institution{Indiana University}
   \city{Bloomington}
   \state{IN}
   \country{USA}}
\email{jiang60@iu.edu}

\author{Fan Chen}
\affiliation{%
   \institution{Indiana University}
   \city{Bloomington}
   \state{IN}
   \country{USA}}
\email{fc7@iu.edu}

\renewcommand{\shortauthors}{}

\begin{abstract}
Large language models (LLMs) increasingly follow neural scaling laws that tie performance gains to rapidly expanding computational budgets, raising concerns about the sustainability of frontier-scale training. Existing carbon-estimation methods largely depend on regression over historical runs and fail to capture critical system-level factors, including hardware heterogeneity, distributed parallelism, communication overhead, and architectural sparsity. We present \textit{CarbonScaling}, a hardware-aware analytical framework for modeling the carbon scaling behavior of frontier LLM training. The framework integrates neural scaling laws, distributed training strategies, accelerator and interconnect modeling, and operational and embodied carbon accounting to estimate feasible hardware configurations and associated emissions. CarbonScaling jointly models tensor, pipeline, data, and expert parallelism while incorporating memory, bandwidth, utilization, and runtime constraints. Experimental validation demonstrates substantially higher fidelity than regression-based baselines and highlights the growing importance of embodied carbon at trillion-parameter scales. Source code: \url{https://github.com/UnchartedRLab/CarbonScaling}.
\end{abstract}



\maketitle

\setlength{\textfloatsep}{10pt plus 2pt minus 4pt} 
\setlength{\floatsep}{8pt plus 2pt minus 2pt} 
\setlength{\intextsep}{10pt plus 2pt minus 2pt}
\setlength{\dbltextfloatsep}{10pt plus 2pt minus 4pt}
\setlength{\abovecaptionskip}{5pt plus 2pt minus 1pt} 
\setlength{\belowcaptionskip}{0pt}

\section{Introduction}

Neural scaling laws~\cite{Ludziejewski:ICML2024,chinchilla} have become a foundational principle in the development of LLMs. Prior work~\cite{Ludziejewski:ICML2024,chinchilla} shows that model performance improves predictably with increasing parameter count, dataset size, and training compute according to power-law relationships. These scaling trends have driven the rapid growth of modern LLMs~\cite{gpt,llama,deepseek,MoonshotAI}, with frontier training runs consuming millions of GPU-hours on large-scale accelerator clusters (e.g., GPT-4 requiring $\sim$50M--60M GPU-hours~\cite{GroesLudvigsen2023}). While scaling laws provide a systematic pathway for improving model capability, they also raise a critical sustainability challenge: understanding how increasing computational scale translates into carbon emissions.

Addressing this challenge is difficult because carbon-efficient scaling depends on substantially more than model size alone. \textit{Operational} carbon emissions~\cite{Varoquaux:FACCT2025} are jointly influenced by hardware generation, accelerator count, distributed training strategy, hardware utilization, interconnect efficiency, and regional electricity carbon intensity. At the same time, \textit{embodied} carbon emissions~\cite{llmcarbon} from accelerator manufacturing are becoming increasingly significant as advanced semiconductor process nodes and High-Bandwidth Memory (HBM) technologies require more carbon-intensive fabrication. Furthermore, frontier LLMs increasingly adopt Mixture-of-Experts (MoE) architectures~\cite{Ludziejewski:ICML2024}, quantization-aware training, and advanced parallelism techniques~\cite{ZERO:SC2020}, which fundamentally alter the relationships among model performance, computation, communication overhead, and energy consumption.

Existing approaches for estimating LLM training carbon emissions are insufficient for capturing these complex interactions. Prior studies~\cite{morand2024green,thompson2023computational,llmcarbon} primarily rely on regression over historical training runs to estimate energy consumption or hardware requirements. However, such methods conflate heterogeneous hardware platforms, training configurations, and architectural designs—for example, combining measurements from a 1B-parameter model trained on H100 GPUs with those from a 2B-parameter model trained on TPUs—thereby limiting their reliability for analyzing future hardware generations or frontier-scale systems. More importantly, regression-based approaches cannot accurately model how distributed training strategies, including tensor, pipeline, data, and expert parallelism, influence training duration, hardware utilization, and communication overhead at scale.

We present \textit{CarbonScaling}, a hardware-aware analytical framework for modeling the carbon scaling behavior of frontier LLM training. Unlike approaches that estimate emissions solely from parameter count or FLOPs, CarbonScaling jointly models neural scaling laws, distributed parallelism, accelerator/interconnect characteristics, and operational and embodied carbon accounting. Given an LLM architecture, hardware platform, and training-time constraint, the framework identifies the minimal feasible device configuration and parallelism strategy that minimize carbon emissions. CarbonScaling enables analysis of key sustainability questions, including the impact of hardware generations, embodied carbon, sparse MoE architectures, and techniques such as quantization on carbon-efficient scaling. Our results show that frontier LLM carbon emissions depend not only on model size, but also on the interaction among hardware scaling, communication overhead, architectural sparsity, and distributed training efficiency. Validation on controlled experiments and reported frontier-scale systems (including DeepSeek and Llama) demonstrates substantially higher fidelity than regression-based baselines.

\begin{table}[t!]
\centering
\footnotesize
\caption{Hardware configurations. (TH: throughput; MCAP: HBM capacity; MBW: HBM bandwidth; link: GPU NVLink or TPU Inter-Chip Interconnect; TDP: thermal design power).}
\label{t:carbon_gpu_spec}
\setlength{\tabcolsep}{2pt}
\begin{tabular}{cccccccc}\toprule
\multirow{2}{*}{Device}  & FP16 TH  & MCAP   & MBW    & link  & TDP    & area     & tech\\
				              & (TFLOPS) & (GB)   & (GB/s) & (GB/s)  & (Watt) & ($mm^2$) & (nm) \\\midrule
V100                  & 119.2    & 32     & 900    & 300     & 250    & 680      & 12    \\
A100                  & 312      & 40     & 1555   & 600     & 400    & 826      & 7    \\
H100                  & 989.4    & 80     & 3352   & 900     & 700    & 814      & 5    \\
B100                  & 1980     & 192    & 8200   & 1.8K    & 700    & 1.6K     & 4NP   \\
TPUv4                 & 275      & 32     & 1200   & 400      & 320    & 600     & 7   \\
TPUv5p                & 459      & 96     & 2765   & 1.2K     & 600    & 426     & 5   \\\bottomrule
\end{tabular}
\vskip -0.2in
\end{table}

\section{Background}
\label{s:back}

\textbf{Neural Scaling Laws}. Neural scaling laws~\cite{Ludziejewski:ICML2024,chinchilla} describe predictable reductions in LLM test loss as model size, dataset size, and training compute increase, typically following power-law relationships. Near-optimal performance~\cite{chinchilla} is achieved by jointly scaling model parameters ($N$), dataset size ($D$), and compute ($C$), where $N \propto D$ and $C \propto N \cdot D$. Recent frontier LLMs increasingly adopt Mixture-of-Experts (MoE) architectures, which activate only a sparse subset of experts per token to increase model capacity without proportional compute growth. Similar to dense models, MoE architectures also exhibit neural scaling behavior~\cite{Ludziejewski:ICML2024}.

\textbf{LLM Training Hardware}. Due to their massive parameter counts, LLMs are trained on distributed multi-node systems~\cite{Narayanan:SC2021}. Nodes~\cite{B200} are connected via high-bandwidth InfiniBand networks, while GPUs within a node communicate through NVLink/NVSwitch interconnects and access local HBM memory. Similar architectures are used in TPU clusters~\cite{schneider2025life}. Although modern accelerators provide extremely high compute throughput, training efficiency is often limited by interconnect and memory bandwidth~\cite{ZERO:SC2020}.

\textbf{Training Parallelism}. To mitigate memory and interconnect bottlenecks, LLM training employs multiple parallelism strategies, including data, tensor, pipeline~\cite{Narayanan:SC2021}, expert~\cite{kim2021scalable}, and ZeRO parallelism~\cite{ZERO:SC2020}. Data parallelism replicates models across devices while sharding data and synchronizing gradients; tensor and pipeline parallelism partition model computation across GPUs; expert parallelism distributes MoE experts with token-routing communication; and ZeRO reduces memory overhead by sharding optimizer states, gradients, and parameters across devices. These strategies strongly influence hardware utilization, communication overhead, training latency, and carbon emissions.

\textbf{LLM Carbon Footprint}. Training frontier LLMs with increasingly large models and datasets results in substantial carbon emissions. The training-related carbon footprint of an LLM consists of two primary components:
\begin{itemize}[leftmargin=*, nosep, topsep=0pt, partopsep=0pt]

\item \textit{Operational carbon} originates from energy consumed during training and is computed from total hardware energy usage, data center power usage effectiveness (PUE), and regional carbon intensity (gCO$_2$e/kWh)~\cite{llmcarbon}. GPU power consumption includes both static and dynamic components~\cite{Kandiah:MICRO2021}: static power reflects leakage and standby energy independent of utilization, while dynamic power scales with hardware activity and utilization.

\item \textit{Embodied carbon} arises from hardware manufacturing~\cite{schneider2025life}. It is estimated using device chip area and carbon-per-area factors~\cite{llmcarbon}, which capture fabrication yield, energy intensity, material sourcing, and chemical emissions. The embodied carbon attributed to training includes GPUs and CPUs, amortized over the ratio between training duration and hardware lifetime.
\end{itemize}

\textbf{Carbon Reduction Techniques}. To reduce the carbon footprint of LLM training, both hardware- and algorithm-level optimizations have been explored. As highlighted in Table~\ref{t:carbon_gpu_spec}, successive GPU and TPU generations (e.g., NVIDIA V100~\cite{V100} to B100~\cite{B200}, TPU v1 to v5p~\cite{schneider2025life}) improve compute throughput and energy efficiency through semiconductor scaling and architectural innovation (Table~\ref{t:carbon_hardware_scaling}), reducing operational carbon emissions~\cite{Akarvardar:IEEE2023}. However, advanced fabrication technologies such as EUV lithography also increase the embodied carbon of accelerators and HBM~\cite{Jones:IEDM2023}. At the algorithm level, techniques such as Quantization-Aware Training (QAT)~\cite{Liu-2024-llm} reduce computation, memory, and communication overhead through low-bit quantization, thereby lowering training energy consumption.

\begin{table}[t!]
\begin{minipage}{0.23\textwidth}
\centering
\centering
\footnotesize
\captionof{table}{Hardware scaling trends~\cite{Akarvardar:IEEE2023}. (TH: throughput; BW: bandwidth).}
\label{t:carbon_hardware_scaling}
\setlength{\tabcolsep}{3pt}
\begin{tabular}{ll}\toprule
hardware              & annual rate \\\midrule
\multirow{2}{*}{core} & TH 1.3; SRAM 1.4; \\
                      & power 1.03; area 1.05 \\
\multirow{2}{*}{HBM}  & BW 1.25; power  \\
                      & 1.03; capacity 1.24\\
NVLink                & BW 1.11\\\bottomrule
\end{tabular}
\end{minipage}
\hfill
\begin{minipage}{0.24\textwidth}
\centering
\includegraphics[width=1.5in]{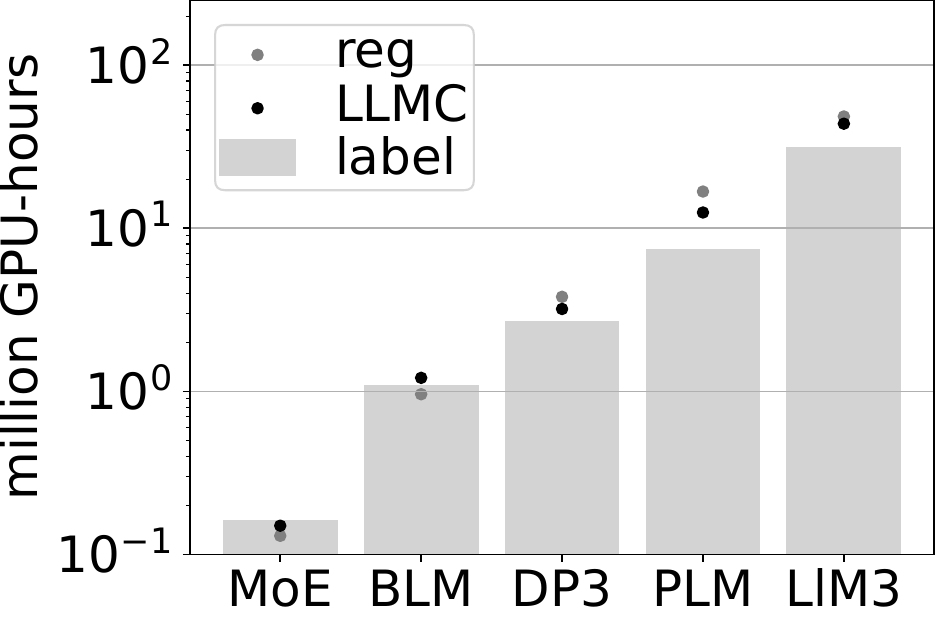}
\captionof{figure}{The inaccuracy of prior regression-based studies.}
\label{f:moti_gpu_num}
\end{minipage}
\vskip -0.1in
\end{table}

\section{Related Work and Motivation}

Neural scaling laws are a major driver of the rapidly increasing carbon footprint of LLM training~\cite{Varoquaux:FACCT2025}. Modeling this behavior requires estimating, for each LLM and hardware platform, the minimal GPU-hours achievable under an optimal parallelism configuration. Prior work~\cite{morand2024green,thompson2023computational} estimates this quantity using regression over historical training runs from heterogeneous hardware platforms. To emulate such approaches (\textit{reg}), we collected minimal GPU-hours from models trained on A100 and H100 GPUs and used them to predict the training requirements of several frontier LLMs, including a 32B MoE model (11B active, MoE), DeepSeek-V3 (DP3)~\cite{deepseek}, BLOOM (BLM)~\cite{luccioni2023estimating}, Llama-3.1 (LlM3)~\cite{llama}, and PaLM (PLM)~\cite{chowdhery2023palm}. The 32B MoE model was trained on 8 H100 GPUs, while training configurations for the remaining models were obtained from prior reports~\cite{deepseek,luccioni2023estimating,llama,chowdhery2023palm}. DP3 was trained on 2048 H800 GPUs, BLM on 384 A100 GPUs, LLM3 consumed 31 million H100 GPU-hours, and PaLM was trained on 6144 TPUv4 accelerators.

As shown in Figure~\ref{f:moti_gpu_num}, ignoring hardware heterogeneity causes \textit{reg} to substantially overestimate minimal GPU-hours relative to measured and reported values (\textit{label}), particularly for large-scale systems such as PaLM. Exhaustively collecting training data across hardware generations is impractical, and regression methods cannot reliably generalize to future hardware without empirical measurements. Similarly, LLMCarbon~\cite{llmcarbon} (\textit{LLMC}), which relies on A100-based regression models~\cite{Narayanan:SC2021}, fails to accurately estimate training requirements for H100 and TPUv4 systems. Both \textit{reg} and \textit{LLMC} ignore critical system-level factors, including hardware configuration, distributed parallelism, batch size, and model bitwidth, limiting their accuracy for frontier-scale LLM training.

\begin{figure}[t!]
\centering
\includegraphics[width=2.6in]{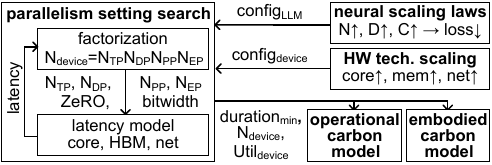}
\caption{The overview of \textit{CarbonScaling}.}
\label{f:carbon_tool_overview}
\vskip -0.1in
\end{figure}

\section{CarbonScaling}

As illustrated in Figure~\ref{f:carbon_tool_overview}, this section presents \textit{CarbonScaling}, which first generates LLM configurations ($\text{config}_\text{LLM}$) by jointly scaling model parameters ($N$), dataset size ($D$), and training compute ($C$). It then incorporates hardware configurations ($\text{config}_\text{device}$), including device frequency, memory capacity, and interconnect bandwidth, together with training strategies such as data ($N_{DP}$), tensor ($N_{TP}$), pipeline ($N_{PP}$), and expert ($N_{EP}$) parallelism, ZeRO sharding, batch sizing, and quantization. For each LLM and hardware configuration, \textit{CarbonScaling} searches over feasible device counts and parallelism settings, enforcing memory and communication constraints to identify the configuration with minimal training duration and maximal utilization. Using the resulting training time, device count, and utilization, \textit{CarbonScaling} estimates total carbon emissions through operational and embodied carbon models.

\textbf{Neural Scaling Laws}. \textit{CarbonScaling} implements dense and MoE neural scaling laws~\cite{Ludziejewski:ICML2024} by varying the model dimension ($d_{model}$). Following modern LLM architectures~\cite{gpt,llama,deepseek}, the feed-forward dimension is set to $d_{ff}=4d_{model}$, while the number of experts scales as $E=8(d_{model}/12288)$. The number of layers follows the fitted scaling law $L=0.402(d_{model})^{0.75}$. The total parameter count is then computed as $N=2d_{model}d_{ff}LE$, with dataset size and training compute given by $D=20N$ and $C=6ND/E$, respectively. The critical batch size scales as 
\begin{equation}
b=E^{0.5}\left(\frac{C}{3\times10^{23}}\right)^{1/6}\frac{2048^2}{len_{seq}}, 
\end{equation}
where $len_{seq}$ is the sequence length. Following~\cite{Ludziejewski:ICML2024}, the training loss is estimated as 
\begin{equation}
\left(18.1+\frac{2.1}{(E/8)^{0.58}}\right)N^{-0.115}+\frac{30.8}{(20N)^{0.147}}+0.47. 
\end{equation}
Setting $E=1$ produces dense LLM configurations. By scaling $d_{model}$, \textit{CarbonScaling} systematically derives the architecture, compute, dataset, and optimization parameters required to model MoE LLM scaling behavior.

\textbf{Hardware Technology Scaling}. \textit{CarbonScaling} instantiates state-of-the-art GPU and TPU configurations using the data summarized in Table~\ref{t:carbon_gpu_spec}. Specifications for NVIDIA GPUs are drawn from~\cite{V100,A100,H100,B200}, while configurations for Google TPUs are adopted from~\cite{schneider2025life}. To model future hardware architectures, \textit{CarbonScaling} extrapolates these specifications using the annual technology-scaling rates described in Table~\ref{t:carbon_hardware_scaling}.

\textbf{Parallelism Setting Search}. Each LLM, training ($\text{config}_{\text{LLM}}$), and hardware ($\text{config}_{\text{device}}$) configuration is evaluated using the search engine in Algorithm~\ref{alg:carbon_search_engine}. The search identifies the minimal device count and parallelism configuration that satisfy a target training duration constraint ($T$) while minimizing training time. The search begins from an idealized device count estimate, $C/(T \cdot \text{PTH}_{device})$, where $C$ is the total training compute and $\text{PTH}_{device}$ is the peak device throughput, and incrementally increases $N_{device}$ until a feasible configuration is found. To reduce communication overhead, expert parallelism is fixed to the number of experts ($N_{EP}=E$). The engine factorizes $N_{device}/N_{EP}$ to enumerate valid combinations of data ($N_{DP}$), tensor ($N_{TP}$), and pipeline ($N_{PP}$) parallelism, while additionally exploring ZeRO stages~\cite{ZERO:SC2020} ($Z\in\{0,1,2,3\}$), where $Z=0$ means no ZeRO. Candidate configurations must satisfy divisibility, memory-capacity, and communication constraints. Microbatching and pipeline interleaving~\cite{Narayanan:SC2021} are configured to improve utilization, while ZeRO sharding reduces memory usage at the cost of additional communication overhead. For each feasible configuration, training duration is estimated using simAI~\cite{Wang:NSDI2025} and Scale-Sim~\cite{Samajdar:ISPASS2020}, which model accelerator cores, memory hierarchies, and interconnect behavior.

\begin{algorithm}[t!]
\footnotesize
\caption{Optimal parallelism search}
\label{alg:carbon_search_engine}
\begin{flushleft}
\textbf{Input:} LLM, training, and hardware configurations\\
\textbf{Output:} minimum training time, device count, utilization
\end{flushleft}
\begin{algorithmic}[1]
\For{$N_{device}\in [N_{ideal}, 2^{30}E]$}
    \State $N_{EP}\gets E$, \quad $duration_{\min}\gets \infty$
    \For{$(N_{TP},N_{DP},N_{PP}) \in \text{factorize}(N_{device}/N_{EP})$}
        \For{$Z \in \{0,1,2,3\}$}
            \State enforce constraints and configure microbatching
            \State apply ZeRO-$Z$ sharding
            \State verify memory and communication overhead
            \State estimate duration via simAI/Scale-Sim
            \State $duration_{\min}\gets \min(duration_{\min}, duration)$
        \EndFor
    \EndFor
    \If{$duration_{\min}<T$}
        \State \Return $duration_{\min}, N_{device}, Util_{device}$
    \EndIf
\EndFor
\end{algorithmic}
\end{algorithm}

\textbf{Carbon Footprint Estimation}. We estimate the total carbon footprint of LLM training by combining operational and embodied carbon models using the training duration, device count, and hardware utilization obtained from the parallelism search engine. 
\begin{itemize}[leftmargin=*, nosep, topsep=0pt, partopsep=0pt]
\item \textit{Operational Carbon} ($CO_{op}$) consists of emissions from computing devices ($CO_{d}$) and supporting system infrastructure ($CO_{o}$). Device-related operational carbon is computed as 
\begin{equation}
CO_{d}=N_{device}\cdot DU \cdot PUE \cdot CI \cdot (P_s + P_d U), 
\end{equation}
where $DU$ is training duration, $PUE$ is data-center power usage effectiveness, $CI$ is regional carbon intensity, $P_s$ and $P_d$ denote static and peak dynamic device power, respectively, and $U$ is device utilization. Supporting infrastructure emissions are estimated as 
\begin{equation}
CO_{o}=N_{sys}\cdot DU \cdot PUE \cdot CI \cdot P_{sys}, 
\label{e:op_energy_all}
\end{equation}
where $N_{sys}$ is the number of server clusters and $P_{sys}$ is the average cluster power consumption. Prior studies~\cite{llmcarbon,schneider2025life} show that $CO_d$ and $CO_o$ are often comparable in magnitude.

\item \textit{Embodied Carbon} ($CO_{emb}$) captures emissions from hardware manufacturing, data-center construction, transportation, and recycling. Chip-fabrication embodied carbon is estimated as 
\begin{equation}
CO_{chip}=\sum_{HW_i \in system}\frac{DU \cdot area_i \cdot CPA_i}{lifetime_i}, 
\end{equation}
where $area_i$, $CPA_i$, and $lifetime_i$ denote the chip area, carbon-per-area factor, and service lifetime of hardware component $HW_i$, respectively. CPA values for major compute and memory components are adopted from~\cite{llmcarbon}. Additional embodied emissions from infrastructure, transportation, and recycling are incorporated following~\cite{schneider2025life}.
\end{itemize}

\begin{figure}[t!]
\centering
\subfigure[GPU-hour.]{\includegraphics[width=1.7in]{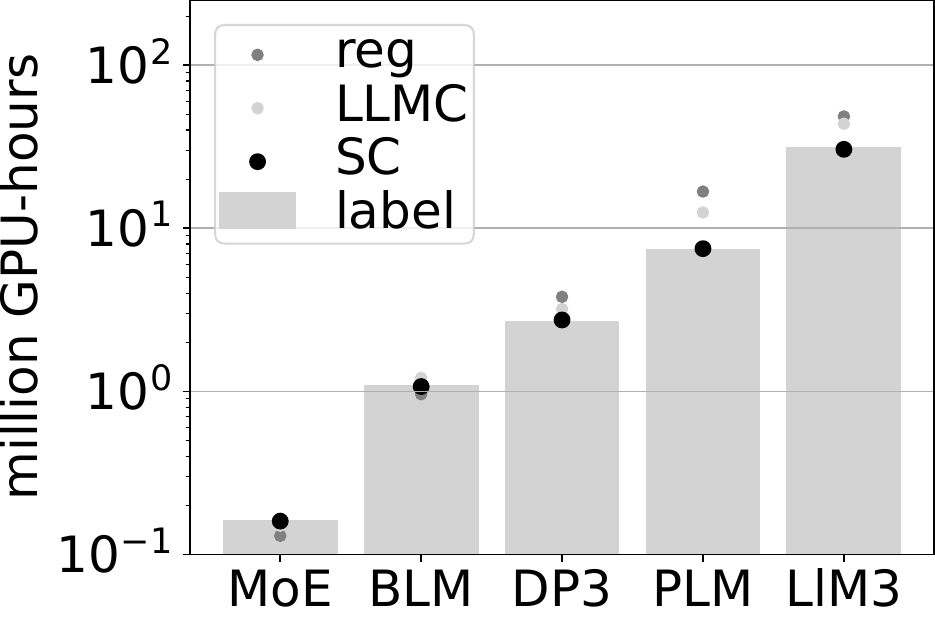}\label{f:calib_gpu_all}}
\subfigure[176B BLOOM Carbon.]{\includegraphics[width=1.6in]{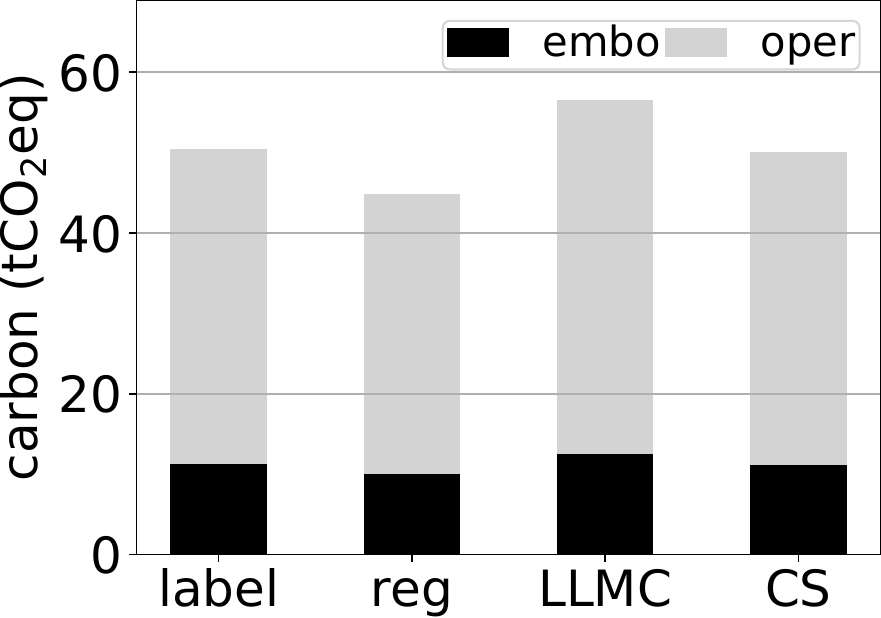}\label{f:calib_gpu_carb}} 
\vskip -0.05in
\caption{The validation of \textit{CarbonScaling}.}
\label{f:carbon_validate_all}
\Description[The validation of \textit{CarbonScaling}.]{The validation of \textit{CarbonScaling}.}
\vskip -0.15in
\end{figure}

\subsection{Validation}

\textit{CarbonScaling} integrates models for neural scaling laws, hardware scaling, device-level latency, parallelism search, and operational and embodied carbon estimation. The neural scaling~\cite{Ludziejewski:ICML2024}, hardware scaling~\cite{Akarvardar:IEEE2023}, GPU/interconnect latency~\cite{Wang:NSDI2025}, and TPU latency~\cite{TPU:Validate} models have been validated in prior work. Here, we focus on validating the parallelism search engine and carbon estimation models.
\begin{itemize}[leftmargin=*, nosep, topsep=0pt, partopsep=0pt]

\item \textit{Parallelism Search Engine}. We evaluate the search engine using a 32B MoE model (11B active, MoE), DeepSeek-V3 (DP3)~\cite{deepseek}, BLOOM (BLM)~\cite{luccioni2023estimating}, Llama-3.1 (LlM3)~\cite{llama}, and PaLM (PLM)~\cite{chowdhery2023palm}, as shown in Figure~\ref{f:calib_gpu_all}. The 32B MoE model was trained on 8 H100 GPUs, while configurations and training durations for the remaining models were collected from prior reports~\cite{deepseek,luccioni2023estimating,llama,chowdhery2023palm}. DP3 used 2048 H800 GPUs, BLM used 384 A100 GPUs, LLM3 consumed 31 million H100 GPU-hours, and PaLM was trained on 6144 TPUv4 accelerators. Measured and reported GPU-hours are denoted as \textit{label}. We compare \textit{CarbonScaling} (\textit{SC}) against a regression baseline (\textit{reg}) trained on A100/H100 data and LLMCarbon~\cite{llmcarbon} (\textit{LLMC}), which relies on A100-based regression models~\cite{Narayanan:SC2021}. Compared with both baselines, \textit{SC} substantially improves GPU-hour estimation accuracy by modeling hardware configuration, distributed parallelism, ZeRO sharding, batch size, and model bitwidth. The coefficient of determination ($R^2$) between \textit{label} and \textit{SC} reaches 99.99\%, compared with only $\sim$98\% for \textit{reg} and \textit{LLMC}.

\item \textit{Operational and Embodied Carbon}. We validate the operational and embodied carbon models in \textit{CarbonScaling} using the 176B BLOOM training run on 384 A100 GPUs. The reported operational energy consumption and embodied carbon are 433,196,kWh and 11.2,tCO$_2$eq, respectively~\cite{luccioni2023estimating}, and are denoted as \textit{label} in Figure~\ref{f:calib_gpu_carb}. For both ground truth and predictions, we assume $PUE=1.2$ and $CI=57,\mathrm{gCO_2e/kWh}$ in Equation~\ref{e:op_energy_all}. We use \textit{CS}, \textit{reg}, and \textit{LLMC} to estimate A100 GPU-hours, which are then converted into operational and embodied carbon using \textit{CarbonScaling}. The total carbon estimated by \textit{CS} deviates from the reported value by only $-0.9\%$, outperforming both \textit{reg} and \textit{LLMC}.
\end{itemize}

\section{Use-Case Studies}

\subsection{Experimental Methodology}

We evaluate \textit{CarbonScaling} by varying the model dimension $d_{model}$ from 1,536 to 24,576, corresponding to LLMs ranging from 0.75B to 24,547B parameters~\cite{MoonshotAI}. The sequence length is fixed at 2K. Following frontier models such as DeepSeek-V3~\cite{deepseek}, configurations below 1.6B parameters are treated as dense, while larger models adopt MoE architectures. We use the GPU and TPU configurations in Table~\ref{t:carbon_gpu_spec} to represent current hardware and apply the projected scaling factors in Table~\ref{t:carbon_hardware_scaling} for future systems. The maximum training duration is set to three months. We assume $PUE=1.1$ and $CI=323,\mathrm{gCO_2e/kWh}$, corresponding to the median value reported for Google data centers~\cite{googlecloud_regioncarbon}, and further evaluate sensitivity to carbon intensity. CPA values are adopted from~\cite{llmcarbon}, with a uniform hardware lifetime of six years.

\begin{figure}[t!]
\centering
\subfigure[Loss v.s. Carbon]{\includegraphics[width=1.6in]{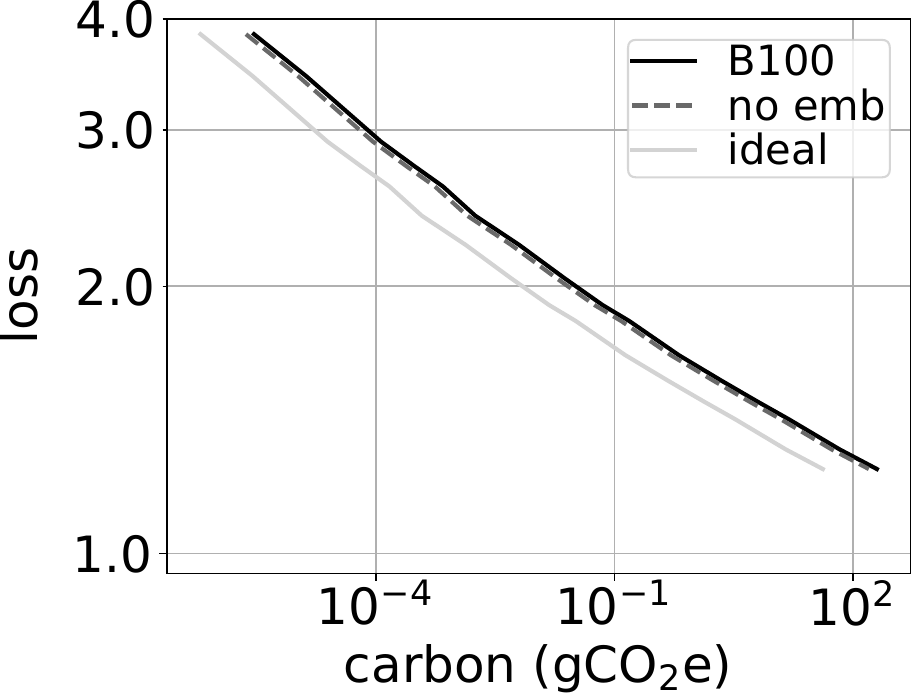}\label{f:carbon_scaling_laws}}
\subfigure[Hardware comparison.]{\includegraphics[width=1.6in]{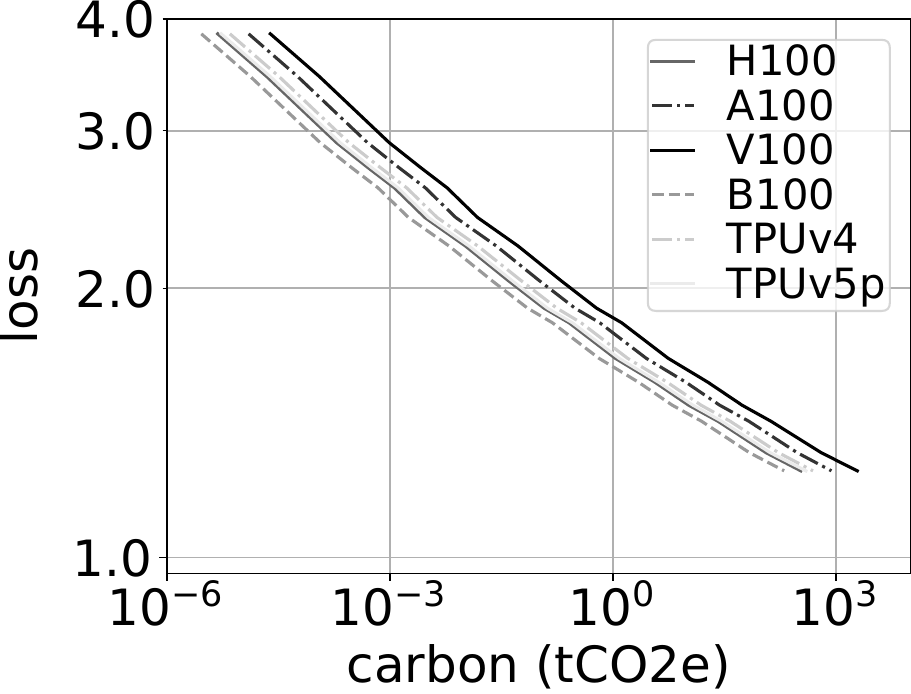}\label{f:carbon_scaling_laws2}} 
\vskip -0.05in
\caption{The neural scaling laws with carbon overhead.}
\label{f:carbon_scale_all}
\Description[The neural scaling laws with carbon overhead.]{The neural scaling laws with carbon overhead.}
\vskip -0.15in
\end{figure}

\subsection{Power-Law Between Accuracy and Carbon}

Figure~\ref{f:carbon_scaling_laws} shows that LLM accuracy follows a power-law relationship with training-related carbon emissions on B100 GPUs. However, practical training incurs substantially higher emissions than the idealized compute-only estimate. The \textit{ideal} curve assumes perfect hardware utilization, no static power, and no embodied carbon, representing the theoretical minimum emissions for the required compute. In contrast, the practical \textit{B100} curve produces $\sim4.7\times$ higher emissions. Removing embodied carbon (\textit{no emb}) shifts the curve closer to the ideal baseline, highlighting the significance of embodied emissions. Prior work~\cite{hardware_scaling} shows that GPU power consumption remains near peak even under reduced utilization—for example, a 37.22\% decrease in utilization results in only a 5.87\% reduction in power—indicating that static power dominates total GPU energy consumption.

\subsection{Impact of Hardware Design and Scaling}

\textbf{Newer GPUs and TPUs}. Newer generations of GPUs and TPUs incur lower training-related carbon emissions than older ones under neural scaling laws. As shown in Figure~\ref{f:carbon_scaling_laws2}, we evaluate carbon-aware neural scaling using NVIDIA V100, A100, H100, and B100 GPUs, as well as Google TPUv4 and TPUv5p. These accelerator generations reflect progressive advances in CMOS process technology and architectural efficiency. The results indicate that, relative to earlier GPU and TPU generations, newer accelerators consistently reduce the carbon footprint required to train LLMs of a given size and target accuracy. Equivalently, for a fixed carbon budget, newer GPUs and TPUs can support the training of larger models with higher accuracy. This underscores the importance of hardware innovation in enabling more carbon-efficient scaling of LLMs. But the incremental carbon savings achieved by successive accelerator generations diminish over time, suggesting decreasing marginal returns from hardware advancements alone. Notably, the training carbon costs of the two TPU generations lie between those of H100 and A100 GPUs and remain substantially higher than those of B100 GPUs, benefiting from more advanced fabrication technologies and architectural designs.

\begin{figure}[t!]
\centering
\subfigure[Carbon per GPU.]{\includegraphics[width=1.6in]{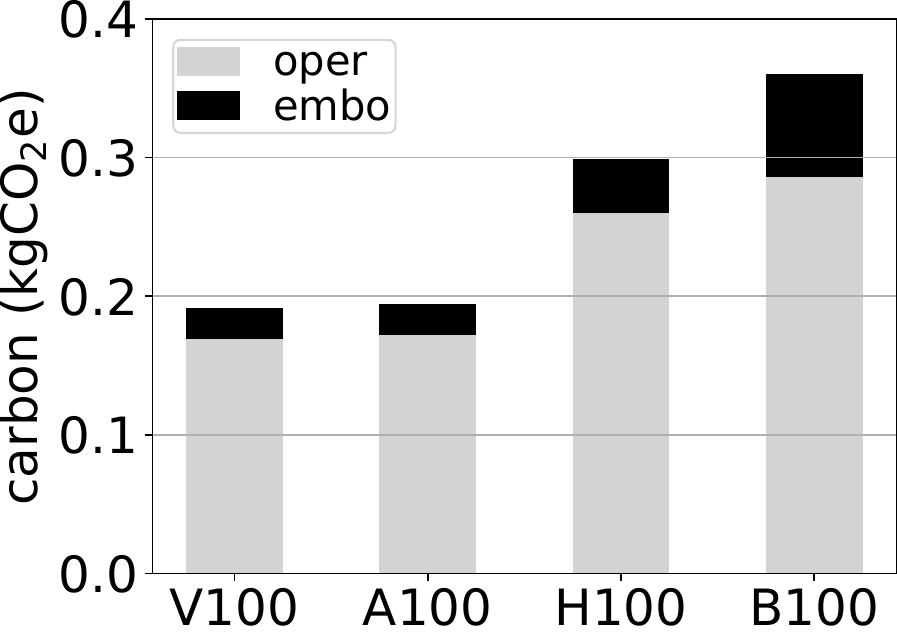}\label{f:carbon_single_comparison}} 
\subfigure[Total carbon.]{\includegraphics[width=1.5in]{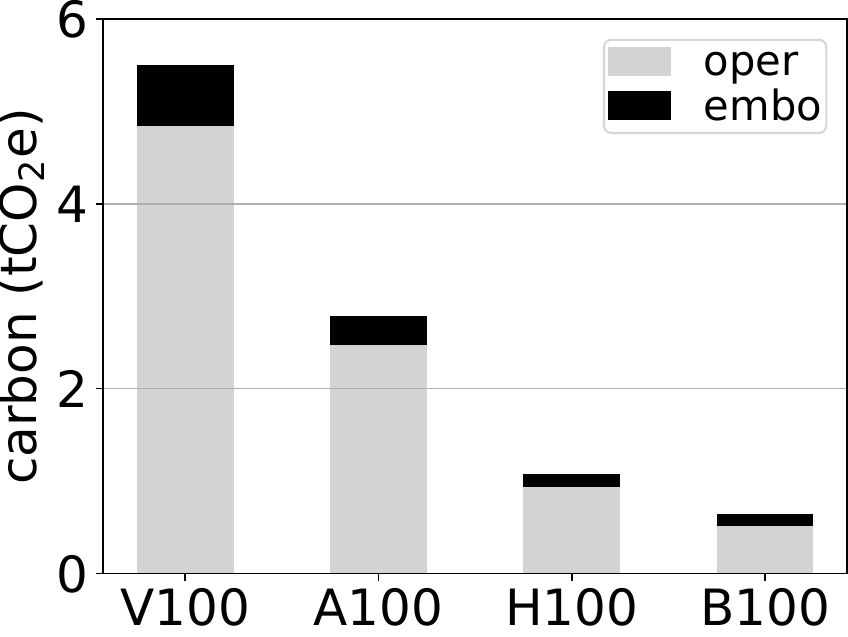}\label{f:carbon_total_comparison}}
\vskip -0.05in
\caption{The training carbon of a 1T-parameter LLM.}
\label{f:carbon_all_comparison}
\Description[the training carbon comparison of a 1T-parameter LLM.]{the training carbon comparison of a 1T-parameter LLM.}
\vskip -0.15in
\end{figure}

\textbf{Further Analysis of Newer GPUs}. Each successive GPU generation delivers higher peak compute throughput and memory bandwidth, often at the cost of increased chip area and higher power consumption. As shown in Figure~\ref{f:carbon_single_comparison}, training a 1T-parameter LLM with newer GPUs incurs higher operational and embodied carbon \emph{per device} due to increased power draw and larger die sizes. However, newer GPUs substantially reduce the total number of devices required for training, as each GPU can accommodate larger model partitions, deliver more compute within a fixed time budget, and lower inter-GPU communication overhead. As shown in Figure~\ref{f:carbon_total_comparison}, despite higher per-device emissions, the overall carbon footprint of training with newer GPUs is significantly lower than that of older generations. In addition, the fraction of embodied carbon increases in newer GPUs. This trend arises because dynamic and leakage energy per operation decrease with smaller process nodes, while fabrication energy rises due to the adoption of EUV lithography and other energy-intensive manufacturing processes~\cite{Jones:IEDM2023}.

\begin{figure}[h!]
\vskip -0.15in
\centering
\subfigure[Hardware scaling.]{\includegraphics[width=1.5in]{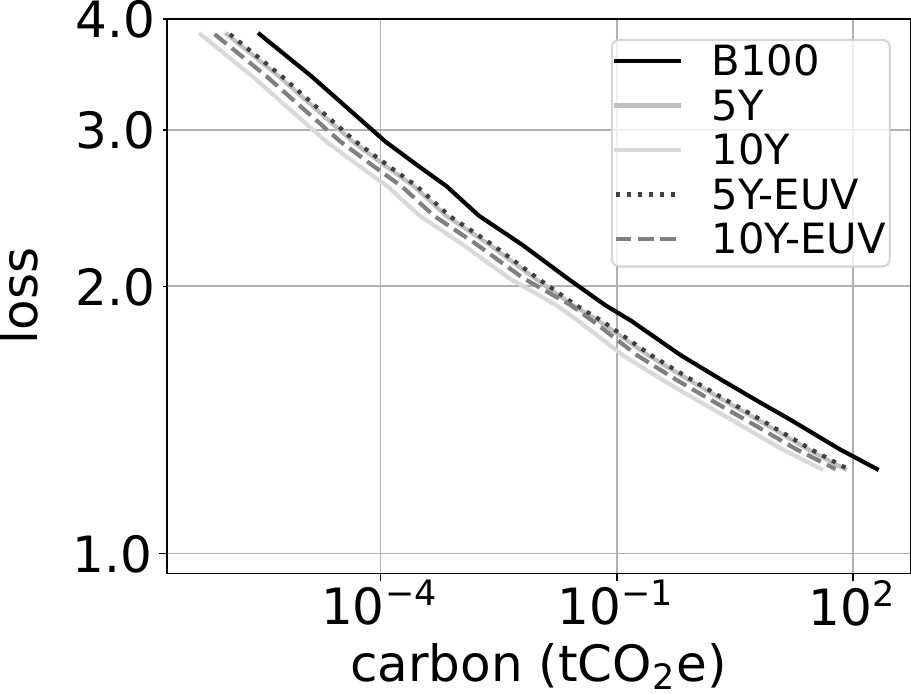}\label{f:carbon_hardware_fut}}
\subfigure[Carbon per GPU.]{\includegraphics[width=1.5in]{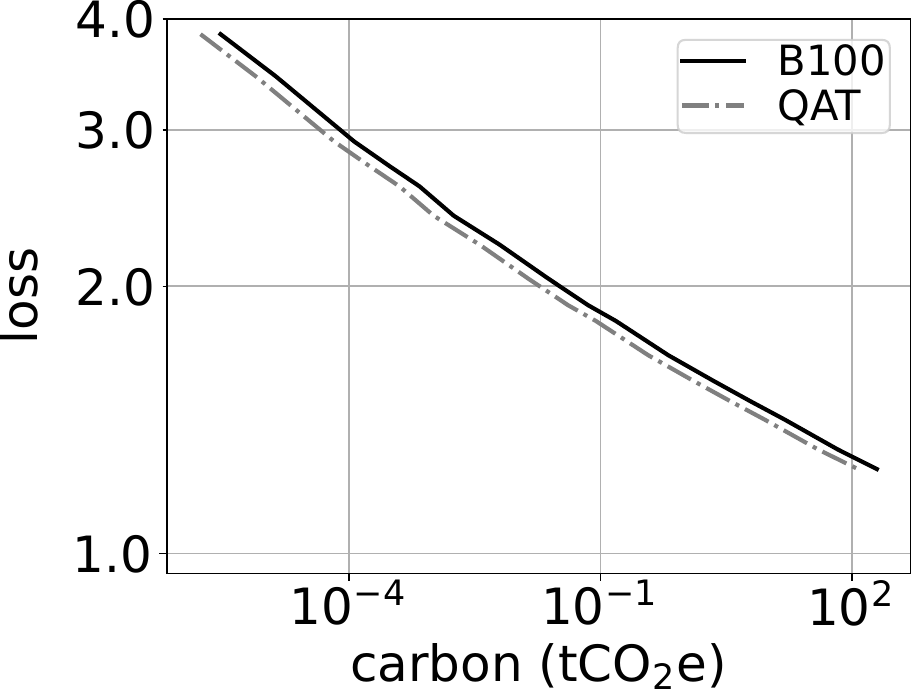}\label{f:carbon_gpu_future}} 
\vskip -0.1in
\caption{Hardware and algorithmic advances.}
\label{f:carb_hard_alg}
\Description[Hardware and algorithmic advances.]{Hardware and algorithmic advances.}
\vskip -0.15in
\end{figure}

\textbf{Hardware Technology Scaling}. Using the B100 configuration and the annual scaling rates in Table~\ref{t:carbon_hardware_scaling}, we project ideal GPU specifications 5 years (5Y) and 10 years (10Y) into the future. As shown in Figure~\ref{f:carbon_hardware_fut}, under ideal assumptions, future GPUs further reduce the training-related carbon footprint of LLMs (by 62\% for 5Y and 83\% for 10Y)—or equivalently improve achievable accuracy under a fixed carbon budget—relative to B100, when increases in embodied carbon from advanced fabrication are ignored. However, the carbon per unit area (CPA) of both logic and HBM memory is projected to increase due to broader adoption of EUV lithography, additional fabrication steps, and higher material and chemical intensity~\cite{Jones:IEDM2023}. When these effects are incorporated by assuming a 10\% annual increase in CPA, the projected 5-year (5Y-EUV) and 10-year (10Y-EUV) curves shift closer to the B100 baseline, indicating diminishing reductions in training-related carbon emissions from future GPU generations. These results suggest that, to sustain efficient carbon scaling, the semiconductor manufacturing industry must increasingly rely on renewable energy sources.

\begin{figure}[t!]
\centering
\subfigure[Carbon scaling w. various CIs.]{\includegraphics[width=1.5in]{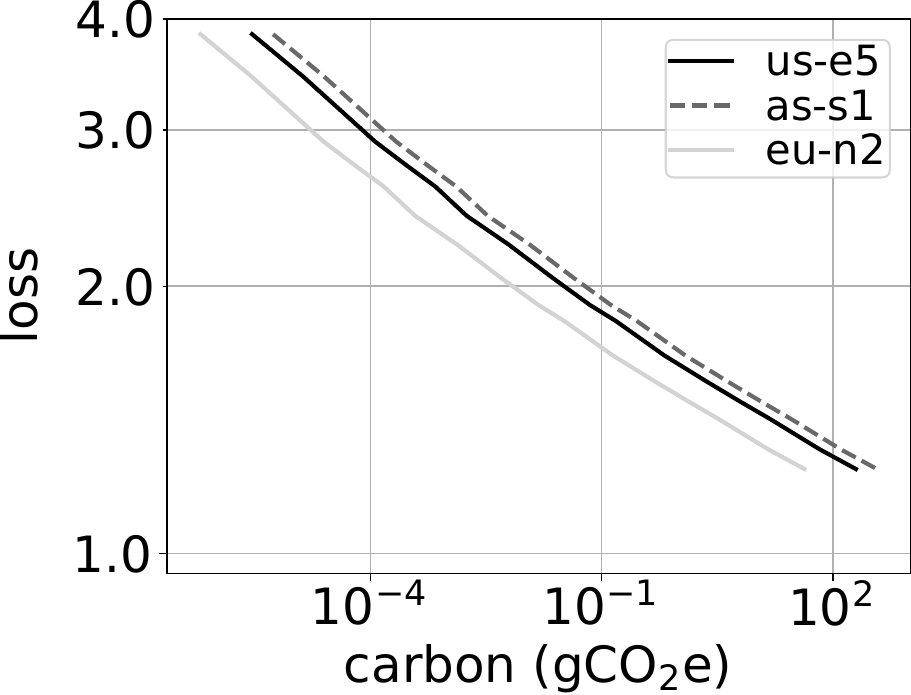}\label{f:carbon_scaling_CI}}
\subfigure[1T-parameter LLM.]{\includegraphics[width=1.6in]{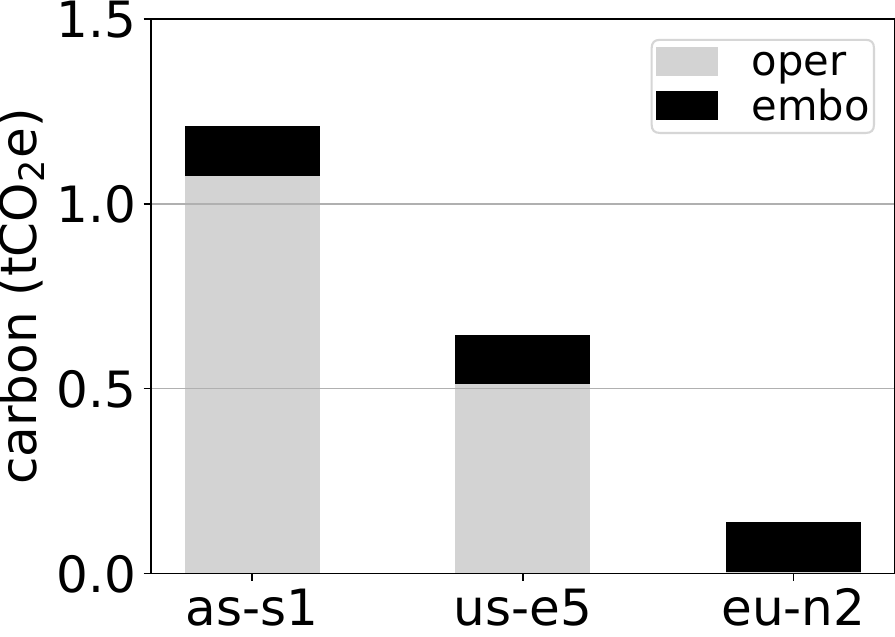}\label{f:carbon_CI_comparison}} 
\caption{The CI sensitivity study using B100 GPUs.}
\label{f:carb_ci_value}
\Description[the CI sensitivity study using B100 GPUs.]{the CI sensitivity study using B100 GPUs.}
\end{figure}

\subsection{Impact of Training Algorithm Advances}

Training algorithm advances can also reduce LLM carbon emissions under neural scaling laws, although their impact is generally smaller than that of hardware scaling. As shown in Figure~\ref{f:carbon_gpu_future}, Quantization-Aware Training (QAT) reduces training-related carbon emissions by approximately 42\% relative to B100. Although smaller than hardware-driven reductions, QAT is easier to deploy across hardware generations and introduces no additional embodied carbon overhead.

\subsection{Sensitivity on Carbon Intensity (CI)}

Figure~\ref{f:carbon_scaling_CI} presents a CI sensitivity study using B100 GPUs. In addition to the medium global CI of 323\,gCO$_2$e/kWh (us-east5), we evaluate the maximum and minimum CI values reported for Google data centers—679\,gCO$_2$e/kWh (asia-south1) and 2.73\,gCO$_2$e/kWh (europe-north2), respectively~\cite{googlecloud_regioncarbon}. Relative to us-east5, the substantially lower CI of europe-north2 reduces training-related carbon emissions by about 80\% across model scales, reflecting its near-complete reliance on renewable energy. As shown in Figure~\ref{f:carbon_CI_comparison}, for a 1T-parameter LLM, the total carbon footprint in us-east5 is $4.6\times$ higher than in europe-north2, where embodied carbon accounts for 97\% of total emissions.

\section{Conclusion}
We present \textit{CarbonScaling}, a hardware-aware framework that extends neural scaling laws to accurately estimate training-related carbon emissions of LLMs. By jointly modeling hardware, parallelism, and embodied and operational carbon, \textit{CarbonScaling} significantly outperforms regression-based methods. Our results reveal a power-law relationship between accuracy and carbon and quantify how hardware advances, algorithmic techniques, and carbon intensity shape carbon-efficient LLM scaling.

\bibliographystyle{ACM-Reference-Format}
\bibliography{scale}

\end{document}